\documentclass[12pt]{article}
\usepackage{aaai}
\usepackage{times}
\usepackage{helvet}
\usepackage{courier}
\usepackage{cite}
\usepackage{times}
\usepackage{latexsym}
\usepackage{algorithm}
\usepackage[noend]{algpseudocode}
\usepackage{amsmath,amssymb}
\usepackage{bm}
\usepackage{picins}
\usepackage{xcolor}
\usepackage{wrapfig}
\usepackage{graphicx}
\usepackage{lettrine}
\usepackage[normalem]{ulem}
\usepackage{todonotes}
\usepackage{setspace}
\usepackage[spaces,hyphens]{url}
\renewcommand{\cite}[1]{(\citeauthor{#1} \citeyear{#1})}
\newcommand{\citealp}[1]{\citeauthor{#1} (\citeyear{#1})}
\newcommand{\citeeg}[1]{\citeauthor{#1}, \citeyear{#1}}

\begin{document}
 \title{Document-editing Assistants and Model-based Reinforcement Learning\\ as a Path to Conversational AI}

\author{Katya Kudashkina\textsuperscript{1,}\textsuperscript{2,3,4},
Patrick M. Pilarski\textsuperscript{3,4},
Richard S. Sutton\textsuperscript{3,4}\\\\
\textsuperscript{1}{University of Guelph}, 
\textsuperscript{2}{Vector Institute for Artificial Intelligence}, \\
\textsuperscript{3}{University of Alberta},\\
\textsuperscript{4}{Alberta Machine Intelligence Institute}\\
Currently under review. Corresponding author: kudashki@ualberta.ca}

\maketitle

\begin{abstract}
Intelligent assistants that follow commands or answer simple questions,
such as Siri and Google search, are among the most economically
important applications of AI.
Future conversational AI assistants promise even greater capabilities
and a better user experience through a deeper understanding of the domain,
the user, or the user's purposes. But what domain and what methods are best
suited to researching and realizing this promise? In this article we argue
for the domain of~\emph{voice document editing} and for the methods of~\emph{model-based reinforcement learning}.
The primary advantages of voice document editing are that the domain is 
tightly scoped and that it provides something for the conversation
to be about (the document) that is delimited and fully accessible to the intelligent assistant.
The advantages of reinforcement learning in general are that its methods are designed
to learn from interaction without explicit instruction and that it
formalizes the purposes of the assistant.
Model-based reinforcement learning is needed in order to 
genuinely understand the domain of discourse
and thereby work efficiently with the user to achieve their goals.
Together, voice document editing and model-based reinforcement learning comprise a promising research direction
for achieving conversational AI.

\end{abstract}

The ambition of AI research is not solely to create intelligent artifacts that 
have the same capabilities as people; we also seek to enhance our intelligence and, in particular, to build intelligent artifacts that assist in our intellectual activities.
Intelligent assistants are a central component of a long history of using computation to improve human activities, dating at least back to the pioneering work of Douglas~\citealp{engelbart1962augmenting}. 
Early examples of intelligent assistants include
sales assistants~\cite{mcdermott1982xsel},
scheduling assistants~\cite{fox1984isis}, intelligent
tutoring systems~\cite{grignetti1975intelligent,anderson1985intelligent},
and intelligent assistants for software development and maintenance~\cite{winograd1973breaking,kaiser1988intelligent}.
More recent examples of intelligent assistants are
e-commerce assistants~\cite{lu_2007_e_commerce_asst},
meeting assistants~\cite{tur2010calo}, and
systems that offer
the intelligent capabilities of modern search engines~\cite{fain2006sponsored,thompson2006google,croft2010search}.
Building intelligent assistants has been positioned as one of the key areas of development in AI~\cite{waters1986kbemacs}.

Today, there are already economically important consumer products based on intelligent assistants.
Intelligent assistants built on voice interaction, such as Amazon Alexa, Google Personal Assistant, Microsoft Cortana,
Apple Siri, Facebook Portal, are among the most important applications of artificial intelligence today.
These {\it voice personal assistants} help people with many daily tasks, 
such as shopping, booking appointments, setting timers, and filtering emails.
The economic value of these systems is created by generating revenue
from selling not only hardware devices, such as smart speakers, but also
voice application features, in the form of sales of digital goods as a one-time purchase, 
subscriptions, or consumables.
The revenue from voice personal assistants alone was forecast to grow 
from \$1.6 billion in 2015 to \$15.8 billion in 2021~\cite{trac}.
It is hard to think of another area in AI that has more immediate social and economic impact.

\section{Conversational and Purposive Assistants}
Today's voice assistants are fairly limited in their conversational
abilities and we look forward to their evolution toward increasing capability.
Smart speakers and voice applications are a result of the foundational research that has come to life in today's consumer products.
These systems can complete simple tasks well:
send and read text messages;
answer basic informational queries;
set timers and calendar entries;
set reminders, make lists, and do basic math calculations;
control Internet-of-Things-enabled devices such as thermostats, lights,
alarms, and locks; 
and tell jokes and stories~\cite{hoy2018alexa}.
Although voice assistants have greatly improved in the last few years, 
when it comes to more complicated routines, such as re-scheduling appointments in a calendar, changing a reservation at a restaurant, or
having a conversation, we are still looking forward to a future where assistants are capable of completing these tasks.
Are today's voice systems ``conversational''?
We say that intelligent assistants are {\it conversational} if they are
able to recognize and respond to input; to generate their own input;
to deal with conversational functions, such as turn taking, feedback, and repair mechanisms;
and to give signals that indicate the state of the conversation~\cite{cassell2000more}.
It is our ambition to achieve a conversational AI assistant that demonstrates these properties,
communicates in free-form language, 
and {\it continuously adapts} to users' changing needs,
the contexts they encounter, and the dynamics of the surroundings.
This conversational AI assistant would understand the domain within which it is assisting and 
provide appropriate support, as well as a pleasant experience for its users.
This kind of assistant has not been developed yet and requires more research.
Conversational AI is a primary goal along the path toward creating intelligent assistants in general.
Achieving conversational AI would lead to even {\it better}
intelligent assistants---intelligent assistants that have a genuine, deeper understanding of their domains
and users, helping people to achieve their goals.

Key to the effectiveness of an intelligent assistant is that it is able to  understand the {\it higher-level goals} of a task when assisting its users.
\citealp{hawkins1968nature} defined a goal or a purpose as a future state that is brought about through
instrumental control of a choice of actions among alternatives.
These higher-level goals or purposes are the reasons for completing a task. 
These purposes motivate and influence smaller intermediate goals.
For example, if a primary goal is flying to San-Francisco, then intermediate goals
can be purchasing flight tickets and packing the luggage.
We use the word ``purpose'' to refer 
to the combination of the high-level context of a task and the user's goals and use the words ``agent''
and ``assistant'' interchangeably.
A user interacting with an intelligent assistant has purposes related to the user's task.
An assistant that understands users' purposes and has its own purposes is a~\textit{purposive intelligent assistant}.
Everyone who has worked on intelligent assistants has recognized the importance of agents'
understanding of purposes, yet an assistant that carries this property has not been developed.
Building learning systems that genuinely understand
purposes has been talked about for decades~\cite{lindgren1968purposive,sun2016intelligent,serban2017hierarchical} and is still ahead on the AI research road map in
fulfilling the promise of assisting people.

Understanding users' purposes is key not only for intelligent assistants in general but, in particular, for conversational AI agents.
Purpose understanding is important for voice assistants in serving their users and
adjusting to the users' unique preferences. 
Imagine a user who wants to have a business meeting with a client
and interacts with a meeting-booking assistant.
In this example, the purpose is a successful business meeting.
The user's initial ask is to book a lunch reservation at a French restaurant for the purpose of the meeting,
which triggers an intermediate goal: to make the reservation.
A meeting-booking assistant that understands the purpose may suggest an Italian restaurant instead.
This is because the assistant has information that the Italian restaurant
is quieter and better suited for business meetings, even though
the assistant realizes that the suggested restaurant is not a requested French type.
The user may accept the intelligent assistant's suggestion, believing that it might
be even~\textit{better} than their initial ask because the suggestion supports the primary purpose, the meeting.
The intelligent assistant's awareness of the purpose is what motivates and drives this scenario.
Another example of such a purposive intelligent assistant is a robotic arm or other manipulation device controlled by a human
user (e.g., a prosthesis or an industrial robot).
Take the case of a human-controlled robotic assistant picking up an object on a chess board or pushing it; both tasks would require different motions or other situation-specific actions from the assistant. If the agent understands the context when assisting with a task, then it would know how to implement a small direct move, like advancing a pawn on the chess board, or an indirect
move, like deploying a knight,
with minimal delays and micromanagement on the user's part.
The assistant creates a smoother user experience
by making better choice of actions when being purposive.

A purposive intelligent assistant improves its capabilities further
when it develops goals or purposes of its own.
This concept is close to an idea where agent-based adjustable autonomy is not prescribed by the user (see~\citeauthor{maheswaran2003adjustable},~\citeyear{maheswaran2003adjustable}).
An example is an iRobot Roomba that cleans a house.
Roomba could set a goal of not hurting itself.  % NOTE: contrast of what it does typically
This intelligent assistant not only seeks and adjusts to the user and their preferences, but also 
develops and sets its own goals and purposes. 
It is important that the user should be aware of the intelligent assistant's goals. 
This awareness would help the user to adjust their preferences, contributing to a more harmonious interaction.
This adjustment between the assistant and the user 
increases the assistant's capabilities
resulting in a better user experience.

\section{The Challenge of Conversation}
Efforts to build voice assistants
that learn purposes are present
not only in modern dialogue systems but go back through four decades
of incremental research and development~\cite{carbonell1970mixed,winograd1971procedures,simmons1972generating,power1974computer,bruce1975belief,walker1978understanding,cohen1978knowing,allen1979plan,pollack1982user,grosz1983team,woods1984natural,finin1986natural,carberry1989plan,moore1989planning,smith1994spoken,kamm1995user}.
Achieving this goal calls for large amounts of computational power,
a great deal of engineering effort to overcome architectural challenges,
and continual human-machine interaction to produce the data.
SHRDLU~\cite{winograd1971procedures} was an early, well-known dialogue system that
responded to instructions and moved objects in a simulated world.
Naively, one could believe that SHRDLU demonstrated understanding of purposes;
however, today it is known that symbolic manipulation systems would fail in the ambiguous
situations humans encounter in real-world settings.

We separate modern dialogue systems
into three categories to illustrate their closeness to purposive intelligent assistants.
The first category is~\textit{entertainment systems} that provide open-domain conversations (see~\citeeg{huang2020challenges}).
These are chatbots and systems that bring a sense of companionship (e.g.,~\citeeg{quarteroni2009designing};~\citeeg{nonaka2012towards};~\citeeg{higashinaka2014towards};~\citeeg{li2016deep};~\citeeg{smith2020can_openDomain};~\citeeg{zhou2020design}),
mostly implemented with sequence-to-sequence
models~\cite{sutskever2014sequence} 
or retrieval-based methods, which select responses from an
existing pre-defined repository.
It is more difficult to develop purposive intelligent assistants
in this category because the user's purpose does not result in a concrete output, and thus is even less firm or clear than in other categories.
The second category is~\textit{text-based instruction systems}
that are primarily represented by text-based games
(e.g.,~\citeeg{he2015deep};~\citeeg{kaplan2017beating};~\citeeg{goyal2019using}).
These dialogue systems are closer to learning purposes.
An example of the work advancing in this direction 
is the use of natural language instructions by~\citealp{goyal2019using}.
The third category is~\textit{task-oriented systems} that help users with particular tasks (see~\citeeg{bing2018_task_oriented}),
such as finding a product, booking a reservation, or call classification (e.g.,~\citeeg{tur2005combining};~\citeeg{bapna2017towards}).
Task-oriented systems category is the closest to our definition of a purposive intelligent assistant,
yet 
the delivery of the task by these systems still depends on pre-designed
slots and templates~\cite{he2015deep,zhao2016towards,lipton2018bbq,liu2018dialogue,goyal2019using,gupta2019casanlu,zhou2019building}
or a handcrafted series of commands that come with if-else conditions and rules.

Modern dialogue systems have rapidly advanced in the last few years, but remain limited in their ability to learn purposes.
The introduction of deep learning techniques through the sequence-to-sequence network by
\citealp{sutskever2014sequence} combined with
the massive amount of data and computational power produced amazing results with the recent system GPT-2~\cite{radford2019language} being a prime example.
But is GPT-2 a step toward purposive intelligent assistants?
We make a simple distinction and clarify that it is a confined language model that
predicts the next word, given all the previous words within some text. 
GPT-2 is an amazing engineering effort that deserves special recognition.
The limitation is that the system's answering capabilities rely on word-by-word prediction,
and not on genuine understanding of a domain or users' goals.
Today's dialogue systems have come a long way yet continue to remain somewhat scripted,
often using a limited number of pre-defined slots and 
canned responses.

Why is it so difficult to get to the goal that would advance today's research community's answers to the challenges of conversational AI and to develop agents that learn purposes? General methods, such as supervised learning, that scale
with increased computation, continue to build knowledge into our agents.
However, this built-in knowledge is not enough when it comes to conversational AI agents assisting users, especially in complicated domains.
One of the biggest problems in conversational AI is the 
{\it limitlessness of domains}, which
leads to high expectations of conversational agents.
In a general conversation, an intelligent assistant is expected to know everything that a human
conversational partner might know and, in some cases, also specialized user-related or world-related information.
For example, the agent is expected to know about relevant aspects or patterns in the wider environment and users' lives, such as what it means to have a schedule.
In other words, the agent is expected to know about the world of its user, and a user can potentially ask the agent anything.
The research community has an ultimate goal to have intelligent assistants
that are able to provide support as effectively as expert human assistants can,
but high expectations with respect to an agent's knowledge about the
world and the users are an obstacle on the path to getting there.

We propose that what is really needed is that the domain is 
{\it tightly scoped and fully accessible} to the intelligent assistant, so that the assistant can fully understand it.
This leads to the question: Is there a domain in which assistants can have focused conversations with their users and be helpful without
knowing everything about the rest of the world but knowing everything about what they have to assist with? We now propose such a domain that allows us to accelerate and advance this field without immediately solving the grand challenge of human-level AI.  
 
\section{Voice Document Editing}
The challenge of genuine understanding in conversational AI requires us to pick a domain that is small enough for an assistant to fully understand.
Document editing is one of the domains that allows an agent to focus a conversation.
Imagine an intelligent assistant that helps create and modify a document via a
free-form language conversation with a user.
This conversation is focused on the document the assistant and the user are authoring.
We call this domain {\it voice document editing}
and propose it as particularly well-suited to develop conversational AI.

The voice document-editing domain fits well into the idea we described earlier: 
learning purposes enables intelligent assistants to become more helpful and powerful.
Document editing assistants could
provide better help if they
could understand users' purposes and allow interactions in a free-form language,
making the interaction process timely and efficient.
They could help create and edit text messages, emails, and other documents on-the-go.
Voice document-editing assistants could perform actions such as deleting and inserting words;
creating and editing itemized lists; changing the order of words, paragraphs, or sentences;
converting one tense to another; or fine-tuning the style.
For example, if a user asks an assistant ``Please move the second paragraph above'', and then says
``Delete the last word in the first sentence'',
then the assistant would know that the first sentence
the user is referring to is in that particular paragraph that was moved by the assistant's previous action.
Such assistants have been emerging for more than three decades~\cite{ades1986voice,douglas1999method,lucas2004method}.

We separate today's dictation systems into two types: the ones that allow voice text modifications~\emph{in addition to dictation} and the ones that do not.
Examples of the latter type include systems  
such as Dragon by Nuance, ListNote, and the Speech Recogniser of iOS~\cite{duffy}. 
Our focus is only on the former type: in these systems, users can write by dictating while walking, cooking, or doing other things, and then {\it edit} the document by using pre-defined commands.
We refer to them as~\emph{voice editing-enabled} systems.
These are dictation software systems such as SMARTedit~\cite{lau2001learning},
Apple Dictation~\cite{gruber2017dictation}, Diction.io~\cite{dictationIO},
Google Docs Voice Editor~\cite{douglas1999method,googleDoc},
and Windows Speech Recognition~\cite{microsoft}.
We now describe how document editing is performed in these systems.

Document editing can be thought of as a manipulation of the manuscript in text blocks.
\citealp{card1980computer} show that, from a cognitive perspective,
document editing is structured into a sequence of almost independent unit tasks. 
The unit tasks are manipulations of selected text blocks,
as suggested by a number of patents related to users' text editing (e.g.,~\citeeg{greyson1997method};~\citeeg{takahashi2001document};~\citeeg{walker1998text}).
In particular, a block of text is first identified by a user.
Next, the block may be moved around, modified, or formatted in place.
A modification operation may include insertion of the new text, or the selected block may be removed completely.
Today's aforementioned voice editing-enabled 
systems~\cite{lau2001learning,gruber2017dictation,dictationIO,douglas1999method,googleDoc,microsoft} 
classify document-modification unit tasks
into text-editing functions representing the types of text-editing
operations people do in their editor of choice.

To demonstrate the suitability of voice document editing for conversational AI, we further look into its advantages.
One advantage of a voice document-editing domain is that it excludes the real-world complexities
that many other assistive systems have.
Consider an intelligent cleaning robot or some other assistive robotic system (e.g.,~\citeeg{dario1996robot};~\citeeg{salichs2019preface}).
These systems have many real-world complexities, such as the
effects of the electronic hardware, materials, sensory systems, surrounding objects, and the variability of sensors.
Voice document editing does not have these dependencies
and its reduced complexity is favorable
for the agent's learning process and for the researchers' experimentation.

An important advantage is that the voice document-editing domain serves as a micro world
with a finite number of clearly defined concepts for the agent to learn.
We refer to this property of the domain as being~\emph{tightly scoped}.
The world is represented by a manuscript being dictated and edited by the user. 
This world is smaller and more manageable than in many other 
real-life applications of conversational AI, such as open domain conversations for chatbots (e.g.,~\citeeg{saleh2019hierarchical}).  
It is easier for the agent to learn in this smaller world because the agent only has to learn about the state of the document and its modifications, both of which are fully accessible to the agent.
At the same time, the agent does not have to understand the content of the document. For example, it is not necessary for the agent to know about history if a user is writing a historical
article, or to know something about medicine if the user is writing a health related article.
The agent is not expected to know information about the user that extends beyond the document editing context: what the user had for breakfast, their religion, other aspects of the user's life, or additional world-related information.
The agent, however, knows about the structure of the document
and can learn about text blocks, such as paragraphs, sentences, and words.
The agent can also know grammatical structure and core organizational components of the document, such as salutations and valedictions when composing an email.
The voice document-editing domain allows the agent to center what a conversation is about---the document itself and its edits.
The finite amount of editing concepts in the fully-accessible document defines fixed bounds for the domain and makes it easier to evaluate the performance.
As a result, relative to other real-life applications of conversational AI, the agent has to learn a smaller number of things, which is favorable for the agent and leads to reasonable expectations.

One way to evaluate our choice of voice document-editing domain is to compare it
to other domains that may carry similar properties of
reduced real-world complexity and being tightly scoped.
We compare our domain to voice image editing and task-oriented systems: 
both are the closest to our definition of the 
purposive intelligent assistant.
We show in which ways voice image editing and task-oriented systems differ
from the domain of our choice, consequently making these domains more challenging
than voice document editing.

A voice image editing assistant has to be able to recognize the content,
which may require the assistant to learn an unlimited number of representations before becoming useful to its users.
Consider a conversational image editing system proposed by~\citealp{manuvinakurike2018conversational}
that is able to recognize voice commands such as ``remove the tree''.
The ability to execute such commands entails
that the system should be able not only to understand the command itself,
but also to have a representation of a tree
to be able to identify a tree in the image that is being edited.
To recognize a tree in an arbitrary image, the agent would have to learn what all possible trees look like. 
Learning about all possible trees requires the assistant to learn an unlimited number of tree representations, which results in the agent having to acquire infinite multi-domain knowledge.
In contrast, in the voice document-editing domain, the agent does not need to understand the content of the document; it only needs to learn how to perform the edits.
For example, the user dictated a phrase ``There was a tree''.
When the user asks to replace  the word ``tree'' with the word ``lake'', the voice editing assistant
does not need to know what the concept of a ``tree'' or a ``lake'' is.
It simply needs to transcribe the word from voice to text and to know the replacement operation, because it already knows the location of words in the text.
This example illustrates that voice image editing is not tightly scoped compared to voice document editing.
While conversational image editing is a suitable domain for incremental dialogue processing, it is more difficult for the agent to become useful to its users in this domain because of the large amount of information it has to learn.

Task-oriented systems also have unlimited concepts for the assistant to learn
despite the focus on a particular user goal. 
Consider a restaurant booking system (e.g.,~\citeeg{Wen_2017Gasic}).
To be a good assistant, the system has to know something about the user's schedule, transportation logistics, and many additional concepts.
For example, the assistant needs to understand that the reservation cannot be made during the time when the user is picking up their children. 
The assistant also needs to know how to select the best location of the restaurant when it comes to transportation logistics.
Scheduling and logistics are only a couple of concept examples that the agent is expected to know about
in addition to knowing the reservation action.
The complexity of the real world leads to a possible unlimited number of concepts
that the agent is expected to know.
This example demonstrates how task-oriented systems may {\it appear} tightly scoped while having unlimited concepts that the assistant has to learn.
In contrast, our choice of voice document-editing domain
makes learning possible because the agent can learn a limited number of concepts
and can still be useful to the user.

Now that we have looked at the advantages of voice document editing, we turn our attention to the current state of such systems. 
Despite enormous effort in this direction, today's voice editing assistive systems remain rudimentary.
When it comes to text modifications using voice, there are limitations: users can format and edit by using only a few pre-defined commands such as
``New line'' in order to start a new line or ``Go to end of paragraph'' when a user wants to move the cursor. 
If the user says slightly modified versions of commands such as ``Let's go to a new line" or ``Move the cursor to the end'' instead of the aforementioned pre-defined commands, then the agent may not perform the right action.
These constraints limit benefits of editing
a document via voice---at the end of the day, a user has to either perform
manual text manipulation using a keyboard or carefully remember
all the commands to manipulate the manuscript via voice.
For example, one of the advanced editors, Google Docs Voice Editor~\cite{googleDoc}, has over a hundred basic commands
that a user would need to memorize or to look up while editing.
Communication to the voice editing assistant in a free-form language is yet to be developed.

An assistant that can communicate to the user in a free-form language
and be helpful to the user is one of the main challenges in conversational AI.
Developing such assistant involves a number of complex elements, that are incredibly challenging in isolation:
from natural language understanding, to acoustic prosody, to natural language generation,
to response generation, to knowledge acquisition~\cite{Mihail2019convAIrecap}.
Voice document editing combines these elements in one domain and thereby opens up research opportunities that could benefit the advancement of conversational AI.
We now propose methods suitable 
for developing voice document-editing assistants.

\section{Reinforcement Learning Assistants}
Reinforcement learning has been pursued as a natural
approach to intelligent assistants~\cite{kozierok1993learning,pollack2002pearl,pineau2003towards}.
Reinforcement learning starts with an agent that is interactive and goal-seeking.
Formally, reinforcement learning is an approach for solving optimal control problems in which a behavior is learned through repeated trial-and-error interactions between a learning system and the world the system operates in.
The learned behavior is called a~\emph{policy}, a learning system is an agent,
and the world the agent operates in is called an~{\it environment}.
In each interaction with the world, the agent takes an action and receives
a~\emph{reward} which can be positive or negative.
The agent aims to maximize the~{\it expected return}, which is the sum of the total rewards in the simplest case~\cite{sutton2018reinforcement}.
Reinforcement learning has not been previously investigated specifically for
voice document editing and
it is the approach that we are going to explore in this article.

The first advantage of reinforcement learning
for intelligent assistants is that
it provides an opportunity for an intuitive
human-computer interaction,
in contrast to more common machine learning formulations of supervised and unsupervised learning.
In particular, an interactive reinforcement learning agent
can directly learn things about its environment with every action
and select future actions according to that knowledge.
This means that agents are not learning from the
input-output pairs that were provided ahead of time, but from direct
experience---{\it online} learning.
Online learning provides a natural opportunity for intuitive interactions,
during which intelligent assistants adapt to users.
Imagine a voice assistant that recommends fun things to do during travel,
similar to the NJFun system~\cite{litman2000njfun,singh2002optimizingNJFun} that provided
users with information about fun things to do in New Jersey.
The assistant can learn about the user's preferences much faster
and provide better recommendations
by extracting a reward signal after each suggestion made and
adjusting its suggestions accordingly.
When this assistant helps a traveler who is interested in music history, it can learn the user's preferences quickly,
based on how satisfied the user was with its previously provided recommendations.
The assistant can tailor the recommended places to music history museums,
music history festivals, music exhibits, and other similar attractions.
This adjustment via online learning between the assistant and the user
improves the assistant's reasoning about its future actions.
The importance of online learning and interaction feedback have appeared in many studies (e.g.,~\citeeg{long_1982_interaction};~\citeeg{pica1986making,pica1987second};
\citeeg{gass_varonis_1994};~\citeeg{bassiri2011interactional};
\citeeg{gavsic2011line};~\citeeg{gavsic2013line};
\citeeg{Ferreira2015_online};~\citeeg{li2017dialogue};
\citeeg{liu2017end}).

The second advantage of reinforcement learning is that users' feedback can be formulated as a reward signal. 
Feedback can be used as a goal-directed signal
of users' satisfaction or dissatisfaction with actions the assistant takes, which is one way to evaluate the assistant (see~\citeeg{jiang2015automatic}).
In the context of dialogue settings, it is expected that users have a way of communicating their feedback to the assistant, either through
voice interaction or physical interaction via a robotic device.
Examples of these interactions include: training assistants with animal-like robot clicker techniques~\cite{kaplan2002robotic}; using a combination of reward signals from a human and an environment~\cite{knox2012reinforcement};
and using a sparse human-delivered training signal, as in the case of adaptable,
intelligent artificial limbs~\cite{pilarski2011online}.
An intelligent assistant maximizing users' satisfaction is a reinforcement learning agent maximizing the expected return.

The third advantage of reinforcement learning is that it allows
\emph{changes} in the agent's knowledge.
These changes are fundamental for learning, which is different from built-in knowledge.
A process when the agent has to change what it knows, rather than knowing whether something is a fact, is defined as~{\it learning}
by~\citealp{selfridge1993gardens} and is considered the most important part of intelligence~\cite{WoodrowH.1946Tatl}.
Knowing is simply factual: one learns a particular kind of knowledge and knows if it is the truth that applies to a particular setting.
Learning, however, leads to understanding.
Understanding is more fluid than drawing knowledge from built-in facts: in understanding, we relate the facts to everything else and can reason about the consequences of our actions.
Reasoning about the outcome 
results in a controlled choice of actions when
presented with alternatives.
As we argued earlier, a controlled choice of actions is a quality of a purposive intelligent assistant that {\it continuously adapts} to users’ changing needs.
Thus, reasoning about the consequences of actions leads to an assistant
that learns how to learn, adapt, and improve by interacting with users (e.g.,~\citeeg{li2017learning_to_learn}).
The most brilliantly-engineered slot-filling systems will not learn
purposes because they cannot learn to avoid repetitive mistakes, nor can they learn to adapt to different users.
In contrast to learning from static datasets, reinforcement learning is more general and allows for these adjustments.
In particular, the structure and other properties of the world are not assumed in reinforcement learning.

For voice document-editing assistants in particular, reinforcement learning is a natural fit.
In reinforcement learning terminology, an agent is a document-editing assistant.
An environment is both a document and the user that interacts with the agent.
The document could be any text of any length, for example, an email, a manuscript, or a text message.
The user could either dictate a new sequence of words or ask
the assistant to modify a block of text: to switch some words or sentences, delete
words, or highlight parts of the text, etc.
The user's communication would be transformed from voice into text
that would then be observed by the agent. 
The agent would react to this communication text and take an action: 
an edit to the document or a request for a clarification from the user.
This action would affect the environment,
potentially resulting in changes to the document in the case of editing actions.
The user would respond in turn to the agent's action by replying with the next request
or by continuing to request edits. 
This response would create a new observation for the agent and this interaction would repeat until the user would be fully satisfied.
The satisfaction of the user would be reflected in a reward signal that the agent receives after every selected action.
We describe such voice document-editing assistants further later in this article, and 
first we address the work that has been already done in applying
reinforcement learning methods to developing conversational AI.

\section{Prior Work Applying Reinforcement Learning to Conversational AI}
\label{RL-in-modern-conv-AI}
Reinforcement learning has been applied to conversational AI in various ways since late 1970s.
Some of earliest works were~\citealp{walker1978understanding,biermann1996composition,levin1997learning,singh2000reinforcement}.
In more recent works,
Ga{\v{s}}i{\'c} et al. (\citeyear{gavsic2011line},~\citeyear{gavsic2013line}) use
reinforcement learning in online settings to directly learn from human interactions using rewards provided by users and to optimize the agent's behavior in reaction to the user.
\citealp{dhingra2016towards} also explore online learning, but in simulated settings.
They train their agent entirely from the feedback that mimics the behavior of real users.
Such simulated settings are not always available for a learning task and
building simulators for dialogue scenarios and tasks is challenging~\cite{cuayahuitl2005human,li2016user}.
To overcome these challenges,~\citealp{zhou2017end} choose to optimize a policy in offline settings using the raw transcripts of the dialogues, while~\citealp{liu2017iterative} take an approach of jointly
optimizing the dialogue agent and the user simulator.
\citealp{xu2018towards} also apply joint modeling of dialogue act selection but use reinforcement learning only to optimize response generation.
A number of others also use reinforcement learning for open-domain dialogue generation(e.g.,~\citeeg{ranzato2015sequence};~\citeeg{li2016deep};~\citeeg{yu2017seqgan};~\citeeg{budzianowski2017sub};~\citeeg{xu2018towards};~\citeeg{jaques2019way}).

One of the big trends in the last five years has been to use
deep reinforcement learning in conversational AI
\cite{he2015deep,cuayahuitl2016deep,shah2016interactive,zhao2016towards,bordes2016learning,budzianowski2017sub,shen2017reasonet,liu2017end,su2017sample,peng2017composite,williams2017hybrid,bing2018_task_oriented,Peng_2018deepdynaQ,tang2018subgoal,weisz2018sample,zhang2018multimodal,mendez2019reinforcement,shin2019happybot,zhaodynamic}, building on the recent success of deep reinforcement learning on games such as Atari, Go, chess and shogi~\cite{mnih2013playing,silver2018general,schrittwieser2019mastering}.
There is a mix of approaches in this body of work.
For example,~\citealp{liu2017end} use reinforcement learning in combination with supervised learning, and then optimize the agent during the interactions with the users.
Shah et al. (\citeyear{shah2016interactive,shah2018building_self_play}) contrast the interpretation of the human feedback as a reward
value~\cite{thomaz2005real,thomaz2006reinforcement,knox2012reinforcement,loftin2014strategy} and propose an interactive reinforcement learning approach in which the user feedback is treated as a
label on the specific action taken by the agent similar to~\citealp{griffith2013policy}.
Reinforcement learning is studied for policy adaptation between domains in multi-domain  settings (e.g.,~\citeeg{gasic-2013-pomdp};
\citeeg{cuayahuitl2017scaling};~\citeeg{gavsic2017dialogue};
\citeeg{rastogi2017scalable};~\citeeg{chen2018policy};~\citeeg{liu2017multi_domain}).
\citealp{serban2017deep} apply reinforcement learning to select from a number of responses produced by an ensemble of so-called response models. 
\citealp{tang2018subgoal} use reinforcement learning to train multi-level policy that allows agents to accomplish subgoals. 
~\citealp{foerster2016learning},~\citealp{sukhbaatar2016learning},~\citealp{lazaridou2016towards},
\citealp{mordatch2018emergence}, and ~\citealp{papangelis2020plato}
apply reinforcement learning to teaching agents to communicate
with each other in multi-agent environments.
These works are a small fraction of the large body of research that implements deep reinforcement learning approaches in dialogue systems.

\section{Model-based Reinforcement Learning}
In the dialogue systems that we have discussed so far, the reinforcement learning agent learns policies and value functions, but not a model of the environment that can be used for planning.
By~\emph{models} of the environment, or~{\it models of the world}, we mean any function that the agent can use to predict how the environment will respond to the agent's actions.
We use the term~\emph{planning} to refer to any computational process that takes a model of the environment as input and produces an improved policy for interacting with the modeled environment.
The idea of augmenting a reinforcement learning agent with a world model that is used for planning
is known as~\emph{model-based reinforcement learning}
\cite{sutton1981adaptive,sutton1985learning,sutton1990integrated,chapman1991input,singh1992_h_dyna,Atkeson_1997_mbrl,wiering2001model,abbeel2007application,sutton2008dyna,ha2018world,holland2018effect,schrittwieser2019mastering}.

Models and planning are helpful.
One advantage of models and planning is that they are useful when the agent faces unfamiliar or novel situations---when the agent may have to consider actions that they
have not experienced or seen before.
Planning can help the agent evaluate possible actions by rolling out {\it hypothetical scenarios} according to the model and then computing their expected future outcomes~\cite{doll2012ubiquity,ha2018world}.
These outcomes can be computed a few steps ahead
and can be thought of as the agent reasoning about the long-term consequences of its actions, similar to how people evaluate the long-term consequences of their decisions.
The agent reasoning about the consequences of its actions
and acting based on the world model's predictions is analogous to
the way a person reasons and acts based on their understanding of the world.
Hypothetical scenarios allow the agent to safely explore the possible consequences of actions.
For example, the agent can use hypothetical scenarios to explore actions that in real-life applications could lead to a costly crash or other disaster~\cite{berkenkamp2017safe}.
Another advantage of world models and planning is that they help accelerate the learning of policies~\cite{freeman2019learning}.

The advantages of world models and planning also arise with intelligent assistants.
World models and planning enable intelligent assistants to leverage
the interaction information as much as possible,
with a minimal amount of prior knowledge and in the absence of external supervision.
Thus, models and planning make it possible for the agent to acquire the full benefits of reinforcement learning in conversational AI,
which in turn would allow us to create a purposive intelligent assistant
that is efficient and useful, can adapt to its users, reason about the consequences of its actions, can control its choice of actions among alternatives, and can learn how the real world works.
Planning has been previously explored in dialogue systems outside of reinforcement learning (e.g.,~\citeeg{stent_2004_planning};~\citeeg{walker2007individual};~\citeeg{Jiang_2019_planning}).

There are a few existing works in conversational AI that already apply 
model-based reinforcement learning (MBRL) methods.
They use a world model to mimic user responses and planning to generate hypothetical experiences that are then used to improve the policy.
\citealp{lewis-etal-2017-deal} introduced dialogue
rollouts, in which planning proceeds by imagining many hypothetical completions of the conversation.
They interleaved reinforcement learning updates with supervised updates.
This work was followed by~\citealp{yarats2017hierarchical} who improved the
effectiveness of long-term planning using rollouts.
\citealp{Peng_2018deepdynaQ} incorporated planning into dialogue policy learning in their deep Dyna-Q framework,
followed by~\citealp{su-etal-2018-discriminativeDynaQ} who proposed to control the quality of hypothetical experiences generated by the
world model in the planning phase.
\citealp{wu2019switch} extended the deep Dyna-Q framework by integrating a switcher that allows to differentiate between a real or hypothetical experience
for policy learning (see also~\citeeg{gao2019neural};~\citeeg{zhang2020recent_advances}).
Subsequently,~\citealp{zhaodynamic} built on the work of~\citealp{Peng_2018deepdynaQ}, using similar world model designs, which is a model-based
approach that is similar to the Dyna architecture~\cite{sutton1990integrated}, where learning and planning are combined and the predictions are improved based on the planning (cf.~\citeeg{lison2013model}).
All these prior works in conversational AI with MBRL are important developments, yet, do not offer a scalable MBRL agent that is fully capable of assisting the user and combines learning and planning in all aspects.
We have begun to explore the combination of learning and planning~\cite{kudashkina2020sampleefficient}.
Below we develop the idea of MBRL and identify its major components, which will help to think further about applying MBRL to the voice document editing domain.

\begin{figure}[b!]
\centering
\includegraphics[scale=0.7]{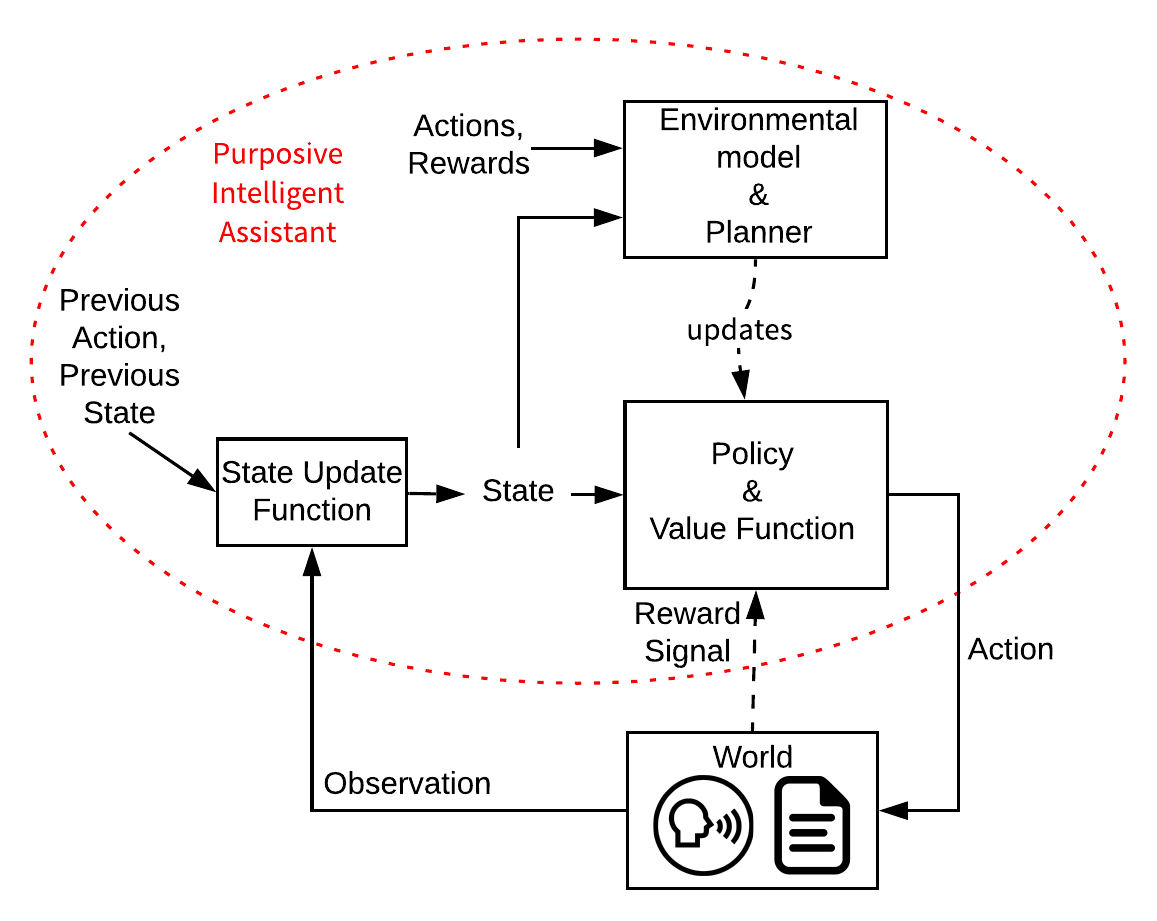}
\caption{A MBRL purposive intelligent assistant with primary components:
an agent state and a state update function; an environmental model; and the agent's policy and value function.
Adopted from~\citealp{sutton2018reinforcement}.}
\label{conceptual}
\end{figure}

There are three primary components of MBRL:
an agent state and its construction; an environmental model;
and a policy, together with a function that estimates the outcome of the agent's actions (Figure~\ref{conceptual}).
The first component of the MBRL agent is
a representation of where it is at the current time---an~\textit{agent state}.
The agent state is an approximation of a minimal environmental state that is the basis of the theory of Markov decision processes.
The agent state is a compact summary of all that has come before, which includes all the histories.
The agent state changes at every time step based on the actions the agent takes and the observed signals from the environment, which we call~\emph{observations}.
The agent state is computed incrementally as a function
of the previous agent state, the most recent observation, and the most recent action.
This function is called a~{\it state update function}.
The state update function is similar to the recursive state update introduced by~\citealp{littman2002predictive}.
The agent state is an essential part of the MBRL architecture and
an input that becomes crucial in the intelligent assistant's decisions.

The second component of MBRL is an environmental model.
As discussed, environmental models are fundamental in making long-term predictions and evaluating the consequences of actions based on those predictions.
\citealp{nortmann2015primary} suggest people's internal models of the world are learned
during the interaction with the world and contain the information that is perceived by their brains.
Similarly, environmental models are learned by the agent during human-computer interaction.
\citealp{forrester1971counterintuitive} describes people's  models of the world as 
the image of the world that humans carry in their head: the concepts and relationship 
between them that are used to represent a real system.
Likewise, environmental models capture the dynamics of the environment and the user-agent interaction within it~\cite{sutton2019MBRL}.
These dynamics are everything that the agent needs to know about the environment.
More precisely, the knowledge of the environment dynamics enables the agent
to predict next states and rewards from the previous state and its actions.
Thus, environmental models help the agent to make informed action choices
by enabling it to compute~\textit{hypothetical} future outcomes of different actions with the modelled environment.
Recall that we refer to this computational process as planning.
Planning is particularly important because it allows the agent to
learn not only from the past actions that resulted in a reward,
but also from hypothetical actions for which the agent has not seen a reward.
\citealp{ng_2003_helicopter} present a successful application of
reinforcement learning that uses a model for autonomous helicopter flight,
saving hours of adjustments and flight testing and preventing unnecessary crashes.
Their model is learned offline prior the agent training. \citealp{ng_2003_helicopter}
start with a human pilot flying the helicopter for several minutes, and then
use the data to fit a model of the helicopter’s dynamics.
Similarly,~\citealp{Coates2017} provide an agent with the pre-trained model and the reward function.

The third component of MBRL includes a policy and a function that estimates the outcome of the agent's actions.
These functions are dependent on states or state-action pairs and are called~\emph{value functions}. 
Policies and value functions are a part of core
reinforcement learning methods that do not include environmental models--- model-free methods.
In simplified terms, value functions compute a numeric value of a given state based on the observed rewards.
These values indicate~\emph{how effective} it is for the agent to be in the given state
or how effective it is to perform a particular action from that state.
Policies and value functions are often approximated by artificial neural networks~\cite{parr2008analysis,sutton2008dyna}.

The policy and the value function, together with the agent state, the state update function,
and the environmental model complete a full MBRL conceptual architecture.
This architecture is general and has no domain-specific components.
This generality could result in a scalable and potentially lasting impact.
The state update function and the environmental model components are critical parts of this
architecture and dependent on each other.
The agent state serves as an input to and an output of the environmental model, 
in addition to being an input for the policy and the 
value function.
Discovering and learning models of environment dynamics
and state update functions are important steps toward learning users' purposes during
interactions while performing assistive tasks.

There is a diversity of open research areas in MBRL.
One area is in the direction of the choice of models (see~\citeeg{chua2018deep}), such as 
probabilistic transition models that use Gaussian processes (e.g.,~\citeeg{kocijan2004gaussian};~\citeeg{deisenroth2013gaussian};~\citeeg{kamthe2018data_eff}), 
nonlinear neural network models (e.g.,~\citeeg{hernandaz1990neural}),
generative models (e.g.,~\citeeg{buesing2018learning_gen_models}),
latent state-space models (e.g.,~\citeeg{wahlstrom2015pixels};~\citeeg{watter2015embed}),
and policy search methods (e.g.,~\citeeg{bagnell2001autonomous};~\citeeg{deisenroth2011pilco};~\citeeg{levine2014learning};~\citeeg{levine2016end}).
Another area of study is on asymptotic performance of model-based methods to match the asymptotic performance of model-free algorithms (e.g.,~\citeeg{chua2018deep}).
The effectiveness of planning methods and data sample efficiency is one more area of active research (e.g.,~\citeeg{hafner2018learning};~\citeeg{kamthe2018data_eff};~\citeeg{kaiser2019_SimPle_model}).
\citealp{silver2017mastering} demonstrate the effectiveness of planning methods in applications such as AlphaGo, offering an algorithm based solely on reinforcement learning, without human data, guidance, or domain knowledge beyond game rules.
Nevertheless, cases of full model-based reinforcement learning similar to~\citealp{silver2017predictron}, in which the environment model is learned from online data and then used for planning, are rare, especially in stochastic domains.
Reduction of errors in learned models that resemble the real environment, data efficiency, asymptotic performance, and a choice of planning methods
are some of the topics at the forefront of MBRL research.

\section{Voice Document Editing with MBRL}
We are at the beginning of a wave of real-world applications of AI that use reinforcement learning methods.
In particular, voice document-editing assistants are one of the purposive intelligent assistants that can be realized using model-based reinforcement learning.
One effort along these lines is what we mentioned before---by~\citealp{kudashkina2020sampleefficient}.
The realization of voice document-editing assistants can be naturally mapped into MBRL components.

The first component of MBRL, the agent state, encapsulates the representation of the document and the history of the conversation and its edits.
The agent's state is 
the current document with its structure, such as paragraphs, sentences, words; what was earlier dictated by the user;
which actions were taken by the agent; and
how satisfied the user was in response to the actions taken.
An observation is represented by a combination of the current document's content and the user's speech.
The observation can be thought of as~\emph{some} information about
what is going on in the environment---the document and the user interacting with the assistant.
The information contained in the observation is partial  because it does not include things such as the user's emotions or anything else outside of the document.
Outside information like users' emotional reactions can be approximated from
additional data such as the time between users' reactions, their tone of voice, or
their facial expressions (e.g., via
techniques such as face valuing by~\citeauthor{veeriah2016face},~\citeyear{veeriah2016face}, or other cues as suggested by~\citeauthor{skantze2016real},~\citeyear{skantze2016real}).

The second component of MBRL, an environmental model, helps to develop
agent's understanding of what happens to the document if it takes an unseen action.
The environmental model captures the dynamics of the user-agent interactions that the agent learns by observing the results of actions taken.
Learning environmental models in voice document-editing domain is possible because of the domain being tightly scoped.
Learning better models leads to improved planning processes that result in better predictions.
Even if the assistant has never heard a command that the user is saying,
by using environmental models and planning, the agent would recognize whether the command is a continuation of a dictation or an edit that the user wants to perform.
When editing requests are recognized, then the agent can plan which editorial action leads to a better outcome and higher users' satisfaction.

In the third component of MBRL, the policy in voice document editing is the agent's behavior in response to user's requests.
This behavior allows the agent to make a choice between continuing to listen to the user
when the user wishes to dictate more, and selecting an editing action to manipulate a text block when the user requests an edit.
The value function in voice document editing represents a numerical value of user's satisfaction.

\section{Remaining Questions}
This article outlined a domain of voice document editing and MBRL methods as particularly well-suited for developing conversational AI.
It is appropriate to discuss some questions that may arise when implementing this proposal, yet implementation details are beyond the scope of this article. One such question is that of how ambitious the agent's learning can be.
Recall that there are many complex elements in conversational AI, such as
natural language understanding, dialogue management, natural language generation, and response generation.
The agent's learning can focus on only one (or on a few) of the elements, while the remaining elements can be treated as a black box; for example, the user's speech can be a black box and can be transcribed by existing tools for speech-to-text conversion (e.g.,~\citeeg{bijl2001speech};~\citeeg{rao2011predictive}).
When not treated as a black box, each of the elements has its own challenges; for example, diversity and coherence in natural language generation element (e.g.,~\citeeg{yarats2017hierarchical};~\citeeg{jang2019bayes};~\citeeg{Shi_2018};~\citeeg{gu2019dialogwae};~\citeeg{wang2019hscjn};~\citeeg{zheng2019persona}).
The agent's ambitiousness in learning that can vary in its complexity with the choices for and within each element of conversational AI.

The agent's learning within one element in conversational AI can be thought of as having three levels of complexity. As an example, consider natural language generation.
The first level of complexity is when the agent's responses being entirely pre-defined by a system designer; this means that no language generation occurs.
The agent's learning then focuses only on the selection of actions to satisfy the user, and the agent's only intelligence is in the learned policy that it follows to select a response.
The model of the environment will learn the dynamics of the interactions and still help the agent with this action selection.
Despite the absence of language generation, this agent can still be useful to its users.
The second level of complexity is when the response generation is treated as a black box and can be supplemented by one of the existing approaches (e.g.,~\citeeg{serban2017deep}) instead of using pre-defined responses as in the first level.
The third level of complexity is when the response generation as an element that is fully learned and this learning becomes a part of the MBRL agent.
The response generation is a demonstration of the agent's ambitiousness in learning within one element of conversational AI.

An ambitious agent is the most desirable; the agent can be fully responsible for learning all the elements of conversational AI, from generating responses to selecting them.
An environmental model will be more complex for this sophisticated agent because
it will have to learn all about the interaction dynamics between the user and the assistant.
These kinds of implementations are often referred to as end-to-end training.
Variations of such implementations can include multiple agents with multiple
models, where each of the agent-model pairs can be responsible for a particular conversational AI element.
Further, there are higher-level models of the world based on extended ways of a temporally abstract behavior, 
which were introduced as~\emph{options} by~\citealp{sutton1998between}.
Higher-level models provide a way of obtaining higher-level planning and reasoning~\cite{sutton2020VectorTalkMBRL}.

Another implementation question is how to train voice document-editing assistants: should they learn online or offline?
Recall that online learning is learning directly from experience, such as from real human-computer interactions.
Offline learning is a traditional approach in supervised learning that relies on pre-constructed datasets.
These datasets can be constructed in a number of ways, including real user-computer interactions, and can be used then to build simulators.

One way an offline dataset can be created is by using the
Mechanical Turk service, a crowdsourcing web service that coordinates the supply and demand of tasks (see~\citeeg{paolacci2010running}).
Using Mechanical Turk, dictation and editing of the resulting documents can be recorded as a dataset,
which is then used to create an environment for voice editing simulations.
As~\citealp{dhingra2016towards} point out, it is common in the dialogue community to use simulated users for this purpose (e.g.,~\citeeg{schatzmann2007agenda};~\citeeg{cuayahuitl2005human};~\citeeg{Asri_2016}).

Another way to obtain offline datasets requires more creative approaches.
Consider the work of~\citealp{feng2019docdial} in which organizational business documents
are used as inputs to generate a conversational offline dataset.
In this conversational dataset, the conversation is based on the organization's workflow.
We encourage the reader to think about similar creative ways of obtaining document-editing datasets.
For example, with some good engineering, the mechanical manipulations of text blocks in manuscripts can be collected and converted into conversational data with the addition of voice commands.

Our preferred method of training is online learning from real interaction data.
Online learning allows the intelligent assistant to adjust to the users during the training process,
and it creates an opportunity for the assistant and the user to build the communication resources developed together during their ongoing interaction (see~\citeeg{pilarski2017communicative}).
Building communication resources can effectively improve collaboration and interaction between the user and the voice-editing assistant.
As~\citealp{mendez2019reinforcement} point out,
a widespread adoption by users of such assistants is limited to the assistants' quality,
which often requires the investment of vast amounts of data.
Thus, even if we were to have access to systems that allow us to experiment with real online data (e.g.,~\citeeg{googleDoc}), then the assistant that learns fully online  might be of lower quality in the beginning of training than one that was pre-trained with offline datasets.
In the past, when a secretary was hired, they were expected to have structural language knowledge.
Similarly, it is reasonable to have such expectations of voice editing assistants and pre-train them
with some preliminary knowledge learned from offline datasets before these assistants can be offered to users and continue to learn online.

\section{Summary and Implications}
In this article, we have proposed that the domain of voice document editing
is particularly well-suited for the development of intelligent assistants that can engage in a conversation.
To make progress in developing useful assistants for conversational AI, these assistants should be purposive.
A natural approach
for developing purposive assistants is reinforcement learning, and, in particular, MBRL.
This approach is well-suited to assistants that learn
and adapt within document editing and general conversational AI settings.
Many aspects of using MBRL remain open areas in AI research, in particular, its use within voice document editing.
Finding solutions for the voice document-editing domain with MBRL and building these systems can provide us with lessons that move us closer to building other systems that genuinely understand the user and learn their purposes.
In this way, a better voice document editing system will also contribute
to the development of 
other assistive systems, moving the research toward the ultimate goal of assistive agents that fully and functionally understand the real world around them. 

The realization of voice document-editing assistants not only serves our objectives of creating a purposive assistant and achieving goals of conversational AI, but also results in an application
that directly benefits society: from improving productivity to benefiting
people with limited typing abilities.

\section*{Funding}
Support for this work was provided in part by the Arrell Food Institute, the University of Alberta, and the Canadian Institute for Advanced Research AI Catalyst Fund Project \#CF-0110.

\section*{Acknowledgments}
We would like to thank Peter Wittek and Joseph Modayil for their useful discussions and feedback.

\newpage

\par\noindent 
\parbox[t]{\linewidth}{
\noindent\parpic{\includegraphics[height=1.5in,width=1in,clip,keepaspectratio]{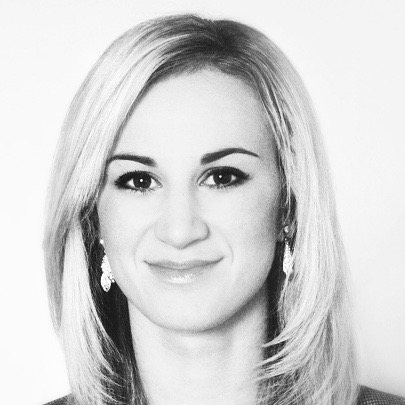}}
\noindent {\bf Katya Kudashkina}\
is pursuing her PhD at the Vector
Institute for Artificial Intelligence and the University of Guelph,
working closely with RLAI Lab at University of Alberta. 
In the past, she has founded two AI startups: UDIO AgTech, and Cultura.
Prior to that she spent over six years at the Canada Pension Plan Investment Board and at IBM.
She studied Engineering in Russia, and then moved to Canada where she completed a degree in Computer Science and then received her MBA at the University of Toronto.}

\par\noindent 
\parbox[t]{\linewidth}{
\noindent\parpic{\includegraphics[height=1.5in,width=1in,clip,
keepaspectratio]{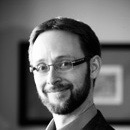}}
\noindent {\bf Dr. Patrick M. Pilarski}\
is a Canada Research Chair in Machine
Intelligence for Rehabilitation at the University of Alberta, an Associate Professor in the Department of Medicine, and a Fellow of the Alberta Machine Intelligence Institute. As part of his research,
Dr. Pilarski explores new machine learning techniques for sensorimotor 
control and prediction, including reinforcement learning methods for human-machine interaction, communication, and user-specific 
device optimization. Dr. Pilarski is the author or co-author of more than 80 peer-reviewed articles, a Senior Member of the IEEE, and has been supported by provincial, national, and international research grants.}

\par\noindent 
\parbox[t]{\linewidth}{
\noindent\parpic{\includegraphics[height=1.5in,width=1in,clip,
keepaspectratio]{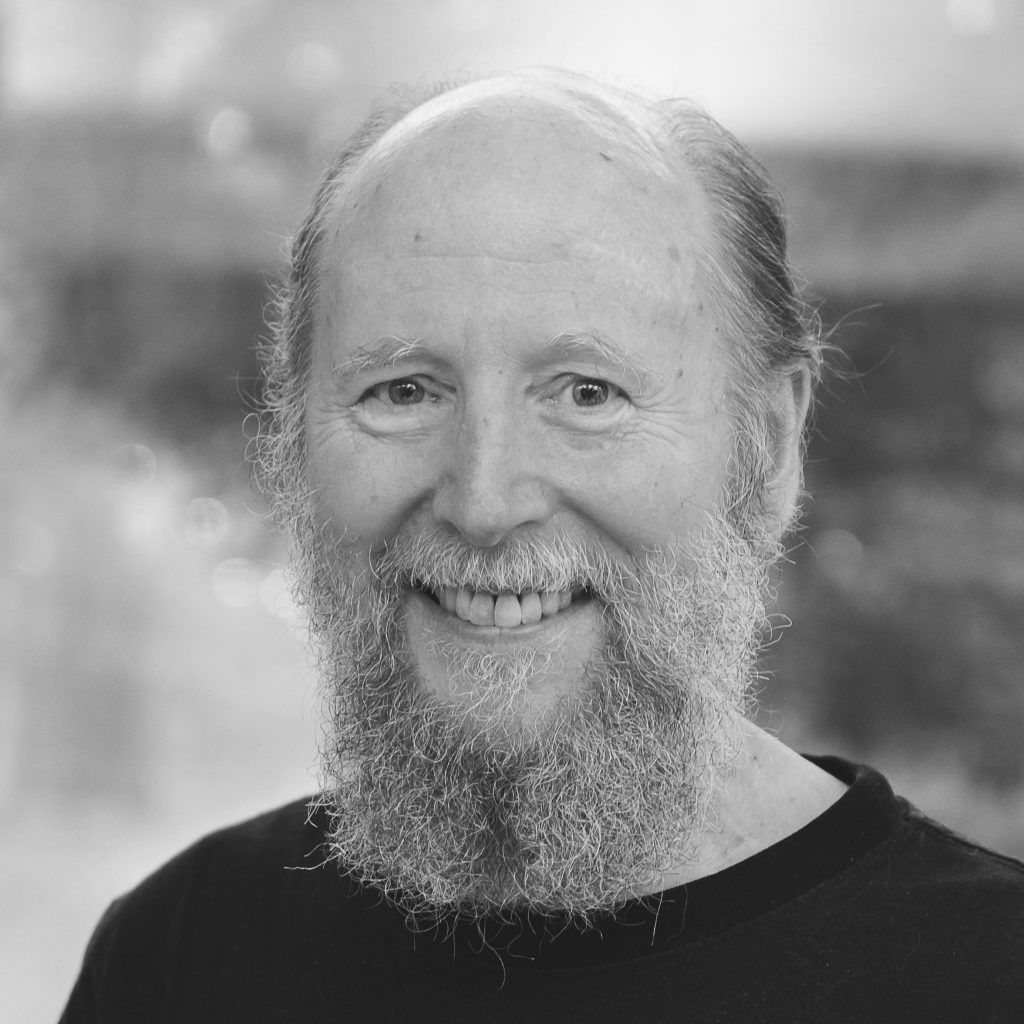}}
\noindent {\bf Richard S. Sutton }\
is a distinguished research scientist at DeepMind
in Edmonton and a professor in the Department of Computing Science
at the University of Alberta. Prior to joining DeepMind in 2017 and
the University of Alberta in 2003, he worked in industry at AT\&T
and GTE Labs, and in academia at the University of Massachusetts.
He received a PhD in computer science from the University of
Massachusetts in 1984 and a BA in psychology from Stanford University in 1978.
He is co-author of the textbook Reinforcement Learning:
An Introduction from MIT Press. He is also a fellow of the
Royal Society of Canada, the Association for the Advancement of
Artificial Intelligence, the Alberta Machine Intelligence Institute, and CIFAR.}
\vspace{4\baselineskip}

\end{document}